%% file: main.tex
\documentclass[10pt]{cai}

\graphicspath{{figs/}}

\usepackage{microtype}
\usepackage{graphicx}
\usepackage{booktabs} 
\usepackage{amsmath} 
\usepackage{subcaption}
\usepackage{multirow}
\usepackage{tabularx}
\usepackage{enumitem}
\usepackage{subcaption}
\usepackage[utf8]{inputenc}
\usepackage[export]{adjustbox}
\usepackage{wrapfig}
\usepackage{algorithm}
\usepackage{algpseudocode}
\usepackage{cuted}
\usepackage{flushend}
\usepackage{hyperref}
\usepackage{standalone}
\usepackage{tikz}
\usetikzlibrary{calc}
\usetikzlibrary{positioning}

\begin{document}

\volumeheader{35}{0}
\begin{center}

  \title{Group and Exclusive Sparse Regularization-based Continual Learning of CNNs}
  \maketitle
  
  \thispagestyle{empty}

  \begin{tabular}{cc}
    Basile Tousside\upstairs{*}, Janis Mohr, Jörg Frochte
   \\[0.25ex]
   {Bochum University of Applied Science, 42579 Heiligenhaus, Germany} \\
  \end{tabular}
  
  \emails{
    \upstairs{*}basile.tousside@hs-bochum.de
    }
  \vspace*{0.2in}
\end{center}

\begin{abstract}
We present a regularization-based approach for continual learning (CL) of fixed capacity convolutional neural
networks (CNN) that does not suffer from the problem of catastrophic forgetting when learning multiple tasks sequentially. This method referred to as Group and Exclusive Sparsity based Continual Learning (GESCL) avoids
forgetting of previous tasks by ensuring the stability of the CNN via a stability  regularization term, which prevents filters detected as important for past tasks to deviate too much when learning a new task. On top of that, GESCL makes the network plastic via a plasticity regularization term that leverage the over-parameterization of CNNs to efficiently sparsify the network and tunes unimportant filters making them relevant for future tasks. Doing so, GESCL deals with significantly less parameters and computation compared to CL approaches that either dynamically expand the network or memorize past tasks' data. Experiments on popular CL vision benchmarks show that GESCL leads to significant improvements over state-of-the-art method in terms of overall CL performance, as measured by classification accuracy as well as in terms of avoiding catastrophic forgetting.
\end{abstract}

\copyrightnotice

\section{Introduction}
\label{sec:intro}
Consider a standard image classification problem, where a CNN is given a data stream explicitly divided into a sequence of $T$ tasks $\{\mathcal{T}_1, ..., \mathcal{T}_T\}$ with $D^t = \{ x^t_n, y^t_n\}_{n=1}^{N^t}$ being the dataset of the $t^{th}$ task and $x^t_n$, $y^t_n$ respectively an instance and label among $N^t$ examples.
In a typical machine learning setup, the model is trained using data from all tasks at once. However, in real-world applications, additional policies have to be considered. In data privacy for example, it is required that old data is deleted after a given time to respect users's privacy concerns. This leads to a situation where data from old tasks is unavailable when learning future tasks. Another example is an online learning setting where data from new tasks arrive on-the-fly.

In such scenarios where the CNN can not access all training data at once, a performance drop occurs on the previous tasks, this phenomenon is referred to as \textit{catastrophic forgetting}~\cite{mccloskey1989catastrophic}. Overcoming catastrophic forgetting while limiting the models capacity, computational cost and memory footprint is the focus of continual learning also termed as lifelong- or sequential learning. 

The main challenge in overcoming catastrophic forgetting is to resolve the \textit{stability-plasticity dilemma}~\cite{abraham2005memory}, which describes the problem of adapting to non-stationary data while at the same time preventing forgetting and is a common issue for both biological and artificial neural networks. 
When an artificial learning model is presented with data in sequence, it needs (i) stability to remember how to solve earlier tasks without re-training on previous data while it also requires (ii) plasticity to acquire new knowledge i.e., to learn new tasks. 

If the model is trained without any immunity against the forgetting of past tasks, therefore focusing on the current task only, it will be very plastic but not stable, meaning that it can learn fast but also forgets quickly. On the other side, if the network mainly focus on being immune against forgetting, which is often achieved by identifying and freezing neurons or weights that were important in learning old tasks, it might lack plasticity i.e., enough capacity to learn future tasks, especially in the case of a fixed-capacity network, which is used in this paper.

To solve the stability-plasticity dilemma in fixed-capacity
networks, typical approaches penalize the deviation of important weights for previous tasks during learning on new tasks. These approaches tend to achieve a high stability but lack plasticity, when the number of tasks becomes large.
In this paper, we propose a dual strategy to train a model on a sequence of tasks. More precisely, during training on task $t$, in addition to constraining the weights of the network that are vital for previous tasks, we simultaneously sparsify the network using a model compression technique similar to~\cite{yoon2017combined}, which leverage the over-parameterization of CNNs in terms of number of filters and feature maps. Therefore, after training on task $t$, our method identifies two sets of filters: (i) those that were crucial for learning tasks up to $t$, which we penalize to not deviate too much via a stability regularizer thus ensuring the stability of the network and (ii) filters that are unimportant in learning task $t$, which are reinitialized and used to learn future tasks, hence ensuring the plasticity of the network. 

Specifically, to sparsify the CNN, we constrain convolutional kernels parameters with a sparsity regularizer, which we denote as plasticity regularizer, consisting of two parts: (i) an \textit{exclusive sparsity} regularization term, which promotes feature discrimination by enforcing the features captured at each layer to be as different as possible, seconded by (ii) a \textit{group sparisity} regularization term, which allows different layers to share important features. Moreover, the degree of both is adapted to allow more feature sharing at lower layers and feature discrimination at top layers, which makes sense for image classification which is the focus of this work.

For the remainder of the paper, we will outline three main contributions. First, we derive a sparsity regularizer combining exclusive- and group sparsity to ensure the plasticity of the CNN in regard to future tasks. Second, we combine the plasticity regularizer with a stability regularizer, which prevents performance deterioration on previous learned tasks. Finally, in the stability regularizer term we propose to adaptively update the importance of each filter based on its post-activation value, such that important filters are not all equal but differ in their degree of importance.

\section{Related Work}
Continual learning has gained much attention in recent
years. In this section, we provide a brief survey of current state of the art approaches to address CL. These approaches are typically categorized in 3 mains groups. We will elaborate more on the first group since the work presented in this paper fall into this category. For an exhaustive survey see~\cite{parisi2019continual}.

\paragraph{\textbf{Regularization-based}}
\vspace{0.4cm}
The first group addresses continual learning by preventing significant changes to the parameters learned for previous tasks~\cite{von2021learning}. The key idea is to define a score indicating the importance of each parameter in the network for learning the previous tasks. When learning a new task, parameters with high importance are discouraged to deviate too much.
As a notable work, Elastic weight consolidation (EWC)~\cite{kirkpatrick2017overcoming} uses a diagonal approximation of the Fisher Information Matrix as a proxy for parameter importance. 
In a similar vein, Memory Aware Synapses (MAS)~\cite{aljundi2018memory} models the importance by the change in the function learned by the network rather than the loss.
These approaches however focus on the stability of the network, therefore largely neglecting its plasticity for learning new tasks. Our work addresses this shortcoming via the introduced plasticity regularizer.
Another notable difference in our approach is that prior methods typically work at a parameter level (individual parameter of a convolution filter for example), whereas ours works on a group of parameters (the entire filter). More recent regularization-based approaches work on a group-level, similar to ours but have some limitations. For instance, HAT~\cite{serra2018overcoming} proposes a per layer binary mask mechanism to capture important filters, but the method requires to know the number of tasks in advance. AGS~\cite{jung2020continual} translates the notion of filter importance to active and inactive filters and implements a network sparsification strategy related to ours, but the method is restricted to group sparsity, which limits its plasticity capability. 

\paragraph{\textbf{Architecture-based}}
\vspace{0.4cm}
In this second group, the capacity of the network is dynamic and additional neurons/filters can be added when a new task arrives. The key idea is to devote different subsets of the final network to each task~\cite{ostapenko2021continual,yoon2019scalable}. 
The part of the network trained on old tasks can remain fixed therefore ensuring the stability whereas additional capacity comes up for new tasks, thus ensuring plasticity.

\paragraph{\textbf{Memory-based}}
\vspace{0.4cm}
This final group stores data from past tasks or data representatives, which is then used during training on new tasks~\cite{henning2021posterior,lopez2017gradient}. 
It is the oldest technique to deal with continual learning and can often easily be combined with methods from other groups.

\section{Method} \label{sec:Methodoly}
Before describing our approach, we briefly introduce the formalism and notation used throughout the paper.

\subsection{Notation}
As mentioned in Section~\ref{sec:intro}, $t$ indexes a task in a sequence of $T$ tasks appearing to a CNN in an online fashion. Each task has a training and test dataset respectively denoted as $D^t_{\text{train}}$ and $D^t_{\text{test}}$. Note that after training on task $t$, $D^t_{\text{train}}$ is entirely discarded and becomes unavailable in the future.
Let $i \in \{0, \dots, L-1\}$ indexes a convolution layer of the CNN model. The parameters for $i$ at task $t$ reside in the kernel tensor
$K_i^t \in {\rm I\!R}^{k_i \times k_i \times c_i \times c_o}$ and the bias term $b_i^t \in {\rm I\!R}^{c_o}$, where $k_i$ is the kernel size, $c_i$ and $c_o$ the number of input and output channels respectively. We therefore denote by $\mathcal{F}_i^t = (K_i^t,b_i^t)$ the parameters of layer $i$ at task $t$.
Furthermore, we use $\mathcal{F}_{i,j}^t \in {\rm I\!R}^{k_i \times k_i \times c_i}$ and $\mathcal{F}_{i+1,:j}^t \in {\rm I\!R}^{k_i \times k_i \times c_i}$ to respectively denote the convolutional kernel of filter $j$ in layer $i$ and the channel $j$ of tensor $\mathcal{F}_{i+1}^t$ that corresponds to filter$\mathcal{F}_{i,j}^t$.  
The set of all filters $\{\mathcal{F}_{i,j}^t\}$ in the network at task $t$ is then given as $\Theta^t$.
In the rest of the paper, we omit the task index $t$ when the context is clear.

\subsection{Defining the Objective Function}
Denoting by $f(\Theta^t)$ the CNN learner at learning task $t$,  
the goal of continual learning is to maximize the performance of $f(\Theta^t)$ at $t$ while minimizing the forgetting on task $\mathcal{T}_1$ to $\mathcal{T}_{t-1}$, both evaluated on the test dataset $D^{t'}_{\text{test}}$ with $ t' \in \{1, \cdots, t\}$. 
The objective function at learning $\mathcal{T}_t$ is then given as,
\begin{equation}
  \mathcal{L}^t(\Theta^t, D^t_{train}) = \frac{1}{N^t} \sum_{n=1}^{N^t}  
  l_n \left( f (x_n^t, \Theta^t), y_n^t\right),
  \label{eq:loss_one_task}
\end{equation}
where $N^t$ is the number of samples in $D^t_{train}$ and $l_n$ is the loss (cross entropy loss) of training instance $n$.
The final training objective for such a continual learning setting after learning all $T$ tasks in sequence can then be written as,
\begin{equation}
  \mathcal{L}(\Theta^T, D_{train}) = \sum_{t=1}^{T}  \mathcal{L}^t(\Theta^t, D^t_{train}).
  \label{eq:loss_all_tasks}
\end{equation}
However, since we are interested in the continual learning setting where $D^t_{train}$ becomes inaccessible after learning task $t$, the objective function in Eq.~(\ref{eq:loss_all_tasks}) can not be directly minimized.
Therefore, when training on $\mathcal{T}_t$, the challenge is to stabilize 
$\sum_{t'=1}^{t-1}  \mathcal{L}^{t'}(\Theta^{t'}, D^{t'}_{train})$ without explicitly measuring it, while at the same time estimating $\Theta^t$ by optimizing Eq.~(\ref{eq:loss_one_task}).

\paragraph{\textbf{Stability Regularizer}}
\vspace{0.4cm}
To satisfy the above mentioned continual learning desiderata when learning a sequence of classification tasks, we constrain for each filter $j$, its convolution kernel parameters learned up to task $t-1$ (denoted $\hat{\mathcal{F}}_{i,j}^{t-1}$) such that their deviation during learning on task $t$ is penalized, which results in stabilizing the performance of the network on previous tasks.
To achieve this, we equip the training objective with a regularizer denoted \textit{stability regularizer} $\mathcal{R}_S(\Theta)$, which forces the difference between $\hat{\mathcal{F}}_{i,j}^{t-1}$ and $\mathcal{F}_{i,j}^{t}$ to be minimal.

Our loss function in Eq~\ref{eq:loss_one_task}, when training on task $t$ can then be rewritten as,

\begin{equation}
  \mathcal{L}^t(\Theta^t, D^t_{train}) = \frac{1}{N^t} \sum_{n=1}^{N^t}  
  l_n ( f (x_n^t, \Theta^t), y_n^t)  + 
  \underbrace{\sum_{\Theta^{t}} 
  \|\mathcal{F}_{i,j} -  \hat{\mathcal{F}}_{i,j}^{t-1}\|_2}_{\mathcal{R}_S(\Theta^t)}.
  \label{eq:stab1}
\end{equation}

The stability regularizer $\mathcal{R}_S(\Theta^t)$ in Eq.~(\ref{eq:stab1}) penalizes the deviation on all filters with the same hardness, treating them as being equally important. Such an assumption is however not true in modern CNN. To address this issue, we define the vector $\hat{\Gamma}^{t-1}$, which captures the importance of each filter in the network in learning tasks up to $t-1$. The importance of a specific filter $j$ in layer $i$ is denoted $\hat{\Gamma}^{t-1}_{\mathcal{F}_{i,j}}$. We will present later on how $\hat{\Gamma}^{t-1}_{\mathcal{F}_{i,j}}$ is computed, for now, let us assume such an importance measure is defined. The binarized version of $\hat{\Gamma}^{t-1}$ is a binary-valued mask vector denoted  $\tilde{\Gamma}^{t-1}$. Each entry in the binary mask $\tilde{\Gamma}^{t-1} \in \{0, 1\}$ indicates whether or not the corresponding filter is important $(1)$ or not $(0)$ at learning tasks up to $\mathcal{T}_{t-1}$. Based on the latter and denoting multiplication with broadcasting by $\odot$, we can perform the operation $\tilde{\Gamma}^{t-1} \odot \Theta^{t-1}$, which divides $\Theta^{t-1}$ into 2 sets $\Theta^{t-1}_+$ and $\Theta^{t-1}_-$ respectively containing important and unimportant filters for learning tasks up to $t-1$.

Incorporating $\hat{\Gamma}^{t-1}_{\mathcal{F}_{i,j}} \in \hat{\Gamma}^{t-1}$ into the stability constraint and constraining only filters, which are important for previous tasks, the stability regularizer in Eq.~(\ref{eq:stab1}) becomes:
\begin{equation}
\mathcal{R}_S(\Theta^t) = 
  \sum_{\Theta_+^{t-1}} 
  \hat{\Gamma}^{t-1}_{\mathcal{F}_{i,j}} 
  \|\mathcal{F}_{i,j} - \hat{\mathcal{F}}_{i,j}^{t-1}\|_2.
  \label{eq:stab2}
\end{equation}

\paragraph{\textbf{Plasticity Regularizer}}
\vspace{0.4cm}
Our loss function as defined in Eq.~(\ref{eq:stab1}) (with the stability regularizer in Eq.~(\ref{eq:stab2})) encourages the network to be stable on previously learned tasks. However, its capacity is not yet optimally used for capturing upcoming knowledge, especially in cases where the number of tasks is large. In experiments, we observed that this loss function provides satisfactory results on small continual learning benchmarks like SVHN~\cite{netzer2011reading}, which consists only of 5 tasks. However for a benchmark like ImageNet-50, which has 25 tasks, the CNN rapidly becomes inefficient at learning future tasks. Consequently, when training on task $\mathcal{T}_{t}$, the challenge is to make the network plastic enough for learning task $\mathcal{T}_{t+1}$ to $\mathcal{T}_{T}$ while maintaining its capacity fixed. Our idea to tackle this issue is inspired by model compression of convolutional neural networks~\cite{liebenwein2019provable}. 
This attractive area of research aims at generating compact CNNs by identifying and removing redundant filters
from an over-parameterized network. The methods proposed to handle this typically start from a CNN that has been trained in a traditional manner, without taking into account such a future compression. In this work, we aim at considering compression during training on each task. Furthermore, instead of discarding unimportant or redundant filters as model compression techniques do, we reinitialize them such that they can be used for learning future tasks. 

Our main idea to account for compression during training is to empower the loss function with a regularizer, denoted \textit{plasticity regularizer}, that will sparsify the CNN such that kernel of redundant filters are zeroed-out. 
The most common regularizer to boost sparsity is the $l_1$-norm $\sum_{\Theta^{t}} \lvert \mathcal{F}_{i,j} \rvert$. However, sparsification using $l_1$-norm usually results in accuracy drop in the context of deep CNN, since it tends to cancel out individual kernel parameters but not the complete filter. The most effective regularizer to sparsify CNNs is the group sparsity regularizer~\cite{wen2016learning}, which deactivates filters entirely at once, thus achieving structured sparsity. Using group sparsity, our plasticity regularizer can take the form:

\begin{equation}
    \mathcal{R}_P(\Theta^t) = \sum_{\Theta_-^{t}} \| \mathcal{F}_{i,j} \|_2.
    \label{eq:plastic1}
\end{equation}

Group sparsity regularizer sparsify a CNN by highly promoting feature sharing among the layers. This is strongly desired in lower layers of a convolution neural network, where feature need to be shared and grouped into most representative geometry. On the other hand, in upper layers, which aim at differentiating between classes, feature discrimination will be a more appropriate perspective. A regularizer that promotes such a feature discrimination has been proposed in~\cite{zhou2010exclusive} and can be defined as $\sum_{\Theta^{t}} \| {\mathcal{F}_{i,j}} \|_1^2 $. In a CNN this results in constraining the convolution filters to be as different as possible from each other. The filters therefore learn disjoint sets of feature, which removes redundancies among them. To allow both feature-sharing (at lower layers) and discriminance (at higher layers) we combine the group sparsity regularizer in Eq.~(\ref{eq:plastic1}) with an exclusive sparsity regularizer. 

As a key ingredient, we apply both sparsity regularizers to the set of filters which has been identified as unimportant for previous tasks, i.e., $\mathcal{F}_{i,j} \in \Theta_-^{t-1}$,
since filters, which are important in learning those tasks are constrained by the stability regularizer to not change much. The resulting plasticity regularizer can then be formulated as:
\begin{equation}
\mathcal{R}_P(\Theta^t) = \underbrace{
   \sum_{\Theta^{t-1}_-} \gamma \| \mathcal{F}_{i,j} \|_2 }_{\text{Group sparsity}} + 
  \underbrace{
  \sum_{\Theta^{t-1}_-} \zeta  \| {\mathcal{F}_{i,j}} \|_1^2,}_{\text{Exclusive sparsity}}
  \label{eq:plastic2}
\end{equation}
with $\gamma = \psi_i$ and $\zeta = \frac{(1-\psi_i)}{2}$, where
$\psi_i = 1 - \frac{i}{L-1}$ adjusts the degree of sharing and discriminating features at each layer, giving more weight to group sparsity in lower layers, whereas exclusive sparsity is dominating in higher layers.

\paragraph{\textbf{Final loss term}}
\vspace{0.4cm}
During training on task $\mathcal{T}_t$ our network ensures both the stability on previous tasks $\mathcal{T}_{t'}, t' \in \{1, \cdots, t-1\}$ and the plasticity for future task $\mathcal{T}_{t''}, t'' \in \{t+1, \cdots, T\}$ by combining Eq.~(\ref{eq:stab1}), (\ref{eq:stab2}) and (\ref{eq:plastic2}) to form the final training objective as:
\begin{multline}
\begin{aligned} 
  \mathcal{L}^t(\Theta^t, D^t_{train}) = \frac{1}{N^t} \sum_{n=1}^{N^t}  
  l_n ( f (x_n^t, \Theta^t), y_n^t)  
  +~
  \mu_S \sum_{\Theta^{t-1}_+} 
  \hat{\Gamma}^{t-1}_{\mathcal{F}_{i,j}} 
  \|\mathcal{F}_{i,j} - \hat{\mathcal{F}}_{i,j}^{t-1}\|_2 \\
  +~
  \mu_P \sum_{\Theta^{t-1}_-} \psi_i \| \mathcal{F}_{i,j} \|_2 
  +~~
 \mu_P \sum_{\Theta^{t-1}_-} \frac{(1-\psi_i)}{2}~   \| {\mathcal{F}_{i,j}} \|_1^2~,
  \label{eq:finalloss}
\end{aligned} 
\end{multline}
where $\mu_S$ and $\mu_P$ are dimensionless strength of stability and plasticity regularizer.

\subsection{Solving the Optimization Problem}
To minimize our regularized learning objective defined in Eq.~(\ref{eq:finalloss}), we use a proximal gradient descent (PGD) approach, which is a broad class of optimization techniques for separable objectives with both smooth and non-smooth terms,
    $\underset{\Theta}{\text{min}} ~g(\Theta) +  h(\Theta)$,
where $g(\Theta)$ is convex differentiable and $h(\Theta)$ is potentially non-smooth~\cite{parikh2014proximal}.
The idea is to first take for each epoch $k \in \{ 0, \cdots, K-1 \}$ a gradient step on $g(\Theta)$ followed by a ``corrective'' proximal gradient step to satisfy $h(\Theta)$:
\begin{equation}
    \Theta^{k+1} \mathrel{\mathop:}= \text{prox}_{\alpha~h}~ 
    \left(\Theta^k - \alpha  \nabla g(\Theta^k)\right),
    \label{eq:proximal_general}
\end{equation}
where $\Theta^k$ is the $k^{th}$ proximal update step and $\text{prox}_{\alpha~g}$ is the proximal operator defined for a function $g$ scaled by a scalar $\alpha>0$ as:
\begin{equation}
    \text{prox}_{\alpha~h}(v) = \arg\min\limits_{\Theta} \left(h(\Theta) +
    \frac{1}{2\alpha} \| \Theta - v \|_2^2 \right).
    \label{eq:proxi_operator}
\end{equation}

\noindent
Let rewrite the training objective in Eq.~(\ref{eq:finalloss}) as: 
\begin{equation}
    \mathcal{L}^t(\Theta^t, D^t_{train})  = \mathcal{L}^t_{\text{CE}}(\Theta) + \mathcal{L}^t_{\text{Reg}}(\Theta),
    \label{eq:loss_compact}
\end{equation}
where $\mathcal{L}^t_{\text{CE}}(\Theta)$ is the ordinary task-specific cross entropy loss on $D^t_{train}$ (first term in Eq.~(\ref{eq:finalloss})) and $\mathcal{L}^t_{Reg}(\Theta)$ is the convex regularization loss, which combines the stability and plasticity regularizer.
The proximal gradient iteration as defined in Eq.~(\ref{eq:proximal_general}) therefore results in minimizing the task specific cross-entropy loss $\mathcal{L}^t_{CE}(\theta)$ only for one epoch, with learning rate $\alpha$, and from the resulting solution applying the proximal operator of the regularizer loss $\mathcal{L}^t_{\text{Reg}}(\Theta)$.

Assuming training on task index $t$, which is omitted in the following for the sake of readability, the gradient descent and proximal gradient step in Eq.~(\ref{eq:proximal_general}) can then be rephrased for our training objective as:
\begin{align} 
\breve{\Theta}^{k+1} \mathrel{\mathop:}= \Theta^k - \alpha  \nabla \mathcal{L}_{\text{CE}} (\Theta^k) \\ 
\Theta^{k+1} \mathrel{\mathop:}= \text{prox}_{\alpha~\mathcal{L}_{\text{Reg}}}~ 
    (\breve{\Theta}^{k+1}).
    \label{eq:proximal_our}
\end{align}

Since the convolution filters are non overlapping, the proximal gradient update in Eq.~(\ref{eq:proximal_our}) can be applied independently for each filter in each layer. Let us now derive this update rule for each filter. 

\paragraph{\textbf{PGD iteration for a single filter}}
\vspace{0.4cm}
Consider the training of a single convolution filter $\mathcal{F}_{i,j}$ at task index $t$, which is omitted in the following for sake of readability. The regularization loss term $\mathcal{L}_{\text{Reg}}(\Theta)$ in Eq.~(\ref{eq:loss_compact}), which consists of the 3 last terms in Eq.~(\ref{eq:finalloss}) can be written as:
\begin{equation}
    \mathcal{L}_{\text{Reg}}(\mathcal{F}_{i,j}) = 
    \begin{cases}
        \mu_S~\hat{\Gamma}^{t-1}_{\mathcal{F}_{i,j}}~ 
    \| \mathcal{F}_{i,j} -\hat{\mathcal{F}}_{i,j}^{t-1} \|_2 
    & \text{if}~\mathcal{F}_{i,j} \in \Theta^{t-1}_+
    \\
    \mu_P \left( \gamma  \|\mathcal{F}_{i,j} \|_2 
    + 
    \zeta \|\mathcal{F}_{i,j} \|_1^2~\right)
    & \text{if}~\mathcal{F}_{i,j} \in \Theta^{t-1}_-.
    \end{cases}
    \label{eq:loss_reg_onefilter}
\end{equation}

\noindent
For small learning rates~$\alpha$, the proximal operator as defined in Eq.~(\ref{eq:proxi_operator}) converges to~\footnote{see Section 1.2 of~\cite{parikh2014proximal} for more details.}: 
\begin{equation}
    \text{prox}_{\alpha~h}(v) = v - \alpha \nabla h(v),
\end{equation}
which let us rephrase Eq.~(\ref{eq:proximal_our}) for a single filter as: 
\begin{equation}
    \mathcal{F}_{i,j}^{k+1} \mathrel{\mathop:}= \breve{\mathcal{F}}_{i,j}^{k+1} - \alpha \nabla \mathcal{L}_{\text{\text{Reg}}}\left(\breve{\mathcal{F}}_{i,j}^{k+1}\right),
    \label{eq:proxi_for_one_layer}
\end{equation}
where $\breve{\mathcal{F}}_{i,j}^{k+1}$ is the result of applying a gradient step via standard SGD optimizers (e.g., Adam). 

The second term on the right hand side of Eq.~(\ref{eq:proxi_for_one_layer}) is the first order derivative of Eq.~(\ref{eq:loss_reg_onefilter}) with respect to $\breve{\mathcal{F}}_{i,j}^{k+1}$, which when derived translates Eq.~(\ref{eq:proxi_for_one_layer}) as follows:
\begin{itemize}
\item If $\mathcal{F}_{i,j} \in \Theta^{t-1}_+$
    \begin{equation} 
    \begin{split}
    \mathcal{F}_{i,j}^{k+1}  \mathrel{\mathop:} & =
    \breve{\mathcal{F}}_{i,j}^{k+1} - \alpha 
    \left(  
    \mu_S~\hat{\Gamma}^{t-1}_{\mathcal{F}_{i,j}}~
    \frac{\breve{\mathcal{F}}_{i,j}^{k+1}-\hat{\mathcal{F}}_{i,j}^{t-1}}
    {\|\breve{\mathcal{F}}_{i,j}^{k+1} - \hat{\mathcal{F}}^{t-1} \|_2}
    \right) \\
     & = (1-\beta)~\breve{\mathcal{F}}_{i,j}^{k+1} + \beta~\hat{\mathcal{F}}_{i,j}^{t-1} \\
     & \text{with}~~~ \beta = \frac{\alpha ~\mu_S~ \hat{\Gamma}^{t-1}_{\mathcal{F}_{i,j}}}{\|\breve{\mathcal{F}}_{i,j}^{k+1} - \hat{\mathcal{F}}_{i,j}^{t-1}\|_2}
    \end{split}
    \label{eq:update_stab_filters}
    \end{equation}
\item If $\mathcal{F}_{i,j} \in \Theta^{t-1}_-$
    \begin{equation} 
        \centering
        \begin{split}
        \mathcal{F}_{i,j}^{k+1}  \mathrel{\mathop:} & =
        \breve{\mathcal{F}}_{i,j}^{k+1} - \alpha~\mu_P
        \bigg(  
        \psi_i~
        \frac{\breve{\mathcal{F}}_{i,j}^{k+1}}{\|\breve{\mathcal{F}}_{i,j}^{k+1}\|_2} 
        + (1-\psi_i)~\|\breve{\mathcal{F}}_{i,j}^{k+1}\|~sign \left(\breve{\mathcal{F}}_{i,j}^{k+1}\right)\bigg)
        \\
         & = (1-\xi)~\breve{\mathcal{F}}_{i,j}^{k+1} - \eta~sign \Big(\breve{\mathcal{F}}_{i,j}^{k+1}\Big)\\
         & \text{with}~~ \xi = \frac{\alpha~ 
        \mu_P~\psi_i}{\|\breve{\mathcal{F}}_{i,j}^{k+1}\|_2}~~~
        \text{and}~~ \eta = \alpha~\mu_P~(1-\psi_i)~\|\breve{\mathcal{F}}_{i,j}^{k+1}\|_1
        \end{split}
        \label{eq:update_plas_filters}
    \end{equation}
\end{itemize}

\noindent
From Eq.~(\ref{eq:update_stab_filters}) we can observe how full stability on the filter (i.e. $\mathcal{F}_{i,j}^{k+1}  = \hat{\mathcal{F}}_{i,j}^{t-1}$) is guaranteed if $\beta = 1$, whereas full plasticity (i.e. $\mathcal{F}_{i,j}^{k+1} =0$) is achieved in Eq.~(\ref{eq:update_plas_filters}) if $\xi = 1$ and $\eta = 0$.
\begin{algorithm}
   \caption{Numerical Optimization Algorithm for task t}
   \label{alg:pgd}
\begin{algorithmic}
   \State {\bfseries Input:} learned parameter $\hat{\theta}^{t-1}$ up to $t-1$, learning rate $\alpha$, regularizer strengh $\mu_s$, $\mu_p$
   \For{each epoch $k$}
   \State $\breve{\theta}^{k+1} = \theta^k - \alpha  \nabla \mathcal{L}_{CE}^t(\theta^k)~~~~~\triangleright$  Update parameters using SGD based on cross entropy loss

   \For{each filter $j$ in layer $i$}

   \State compute $\psi_i ~~~~~\triangleright$ Adjust degree of sharing and discriminating features.
   
   \State Update $\theta^{k+1}_{\mathcal{F}_{i,j}}$ by computing (\ref{eq:update_stab_filters}) or (\ref{eq:update_plas_filters})

   \EndFor
   \EndFor
\end{algorithmic}
\end{algorithm}

\subsection{Filters Pruning and Reinitialization}
We now present the computation of filters importance, which plays a crucial role in our approach.  

\paragraph{\textbf{Filter importance $ \Gamma^{t}_{\mathcal{F}_{i,j}}$ at learning task $t$.}} 
\vspace{0.4cm}
Once the CNN is trained on task $t$ and the plasticity regularizer has ensured network sparsification, we measure the importance of each filter in the CNN at learning task $t$. The importance $\Gamma^{t}_{\mathcal{F}_{i,j}}$ of a filter $j$ in layer $i$ at learning t is quantified as the average standard deviation of its post-activation value across all training samples of task $t$. More specifically, for each filter $\mathcal{F}_{i,j}^t$, we compute the standard deviation $\sigma_{\mathcal{F}_{i,j}}^t \in {\rm I\!R}^{H_o \times W_o}$ of its output activation $a_{\mathcal{F}_{i,j}} \in {\rm I\!R}^{H_o \times W_o}$ across all $N^t$ training samples:
\begin{equation}
    \sigma_{\mathcal{F}_{i,j}}^t = \sqrt{\frac{1}{N^t}~\sum_{n=1}^{N^t}
    \left(a_{\mathcal{F}_{i,j}} (x_n^t) - \overline{a_{\mathcal{F}_{i,j}}} \right)^2},
    \label{eq:filter_importance_std}
\end{equation}

\begin{figure}
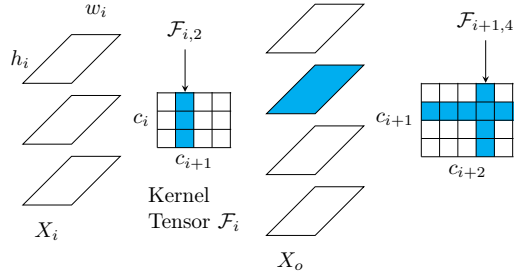

\centering
\includestandalone[width=0.5\linewidth]{figs/filters}
\caption{Processing unimportant filters. Filter $2$ in layer $i$ ($\mathcal{F}_{i,2}$) is unimportant. Consequently, its kernel weights get reinitialized and channel $2$ of all filters in layer $i+1$ ($\mathcal{F}_{i+1,:2}$) is zeroed. The same applies to $\mathcal{F}_{i,4}$, which is also unimportant.}
\label{fig:filter_reinit}
\end{figure}

\noindent
where $a_{\mathcal{F}_{i,j}} (x_n^t)$ is the Rectified Linear Unit (ReLU) activation value of filter $\mathcal{F}_{i,j}$ for the training instance $x_n^t \in D^t_{\text{train}}$ and
$\overline{a_{\mathcal{F}_{i,j}}}$ is the average activation. The importance $\Gamma^{t}_{\mathcal{F}_{i,j}} \in {\rm I\!R}$ of $\mathcal{F}_{i,j}^t$ is then computed by averaging the element of the 2D tensor $\sigma_{\mathcal{F}_{i,j}}^t$ as:
\begin{equation}
    \Gamma^{t}_{\mathcal{F}_{i,j}} = \frac{1}{H_o \times W_o} \sum_{q \in \sigma_{\mathcal{F}_{i,j}}^t} q
    \label{eq:filter_importance_mean_std}
\end{equation}

\paragraph{\textbf{Intuition behind $\Gamma^{t}_{\mathcal{F}_{i,j}}$.}}
\vspace{0.4cm}
The intuition behind the filter importance measure as presented in Eq.~(\ref{eq:filter_importance_mean_std}) with Eq.~(\ref{eq:filter_importance_std}) is based on the observation that one of the reasons for the popularity of the ReLU activation is that it induces a sparsity in activation, which allows convolutional layers to act as feature detectors~\cite{molchanov2016pruning}. Hence, one can reasonably assume that if the output activation value of a filter is small, then the feature detected by this filter (and consequently the filter itself) is not important for learning the current task.

\paragraph{\textbf{Filter importance $\hat{\Gamma}^t_{\mathcal{F}_{i,j}}$ at learning task $\mathcal{T}_1$ up to  $\mathcal{T}_{t}$.}}
\vspace{0.4cm}
Once the importance $\Gamma^{t}_{\mathcal{F}_{i,j}}$ of a filter ${\mathcal{F}_{i,j}}$ in learning task $t$ only, is computed, we need to account for the importance of that filter in learning task $\mathcal{T}_1$ to $\mathcal{T}_{t-1}$ such that it remains stable at performing those tasks. For this purpose, we compute the importance $\hat{\Gamma}^{t}_{\mathcal{F}_{i,j}}$ of filter $\mathcal{F}_{i,j}$ for learning task $\mathcal{T}_1$ to $\mathcal{T}_t$ as:
\begin{equation}
    \hat{\Gamma}^{t}_{\mathcal{F}_{i,j}} := 
    \nu ~ \hat{\Gamma}^{t-1}_{\mathcal{F}_{i,j}} +  \Gamma^{t}_{\mathcal{F}_{i,j}},
   \label{eq:filter_update}
\end{equation}
where $\hat{\Gamma}^{t-1}_{\mathcal{F}_{i,j}}$ is the importance of $\mathcal{F}_{i,j}$ at learning tasks $\mathcal{T}_1$ to $\mathcal{T}_{t-1}$ and $\nu$ is a hyperparameter balancing the filter importance before and after training on task $t$.

\paragraph{\textbf{Processing unimportant filters.}}
\vspace{0.4cm}

For filters with a score $\hat{\Gamma}^t_{\mathcal{F}_{i,j}} = 0$, i.e. filters $\mathcal{F}^t_{i,j} \in \Theta_-^t$, which have been identified as unimportant for learning task up to $t$, we perform the following two actions, which are illustrated in Figure~\ref{fig:filter_reinit}:

\begin{itemize}
    \item Kernel of $\mathcal{F}^t_{i,j} \in \Theta_-^t$ are randomly re-initialized. This allows them to be active and ready to be used by the training procedure for learning a new task.
    \item Kernel of channel $j$ of tensor $\mathcal{F}_{i+1}$ that corresponds to filter $\mathcal{F}^t_{i,j}$ are set to zero, this prevents them from negatively affecting inference on task $t$.
\end{itemize}

\begin{figure*}[ht]
    \centering
    \begin{subfigure}[b]{0.32\textwidth}
        \includegraphics[width=\textwidth]{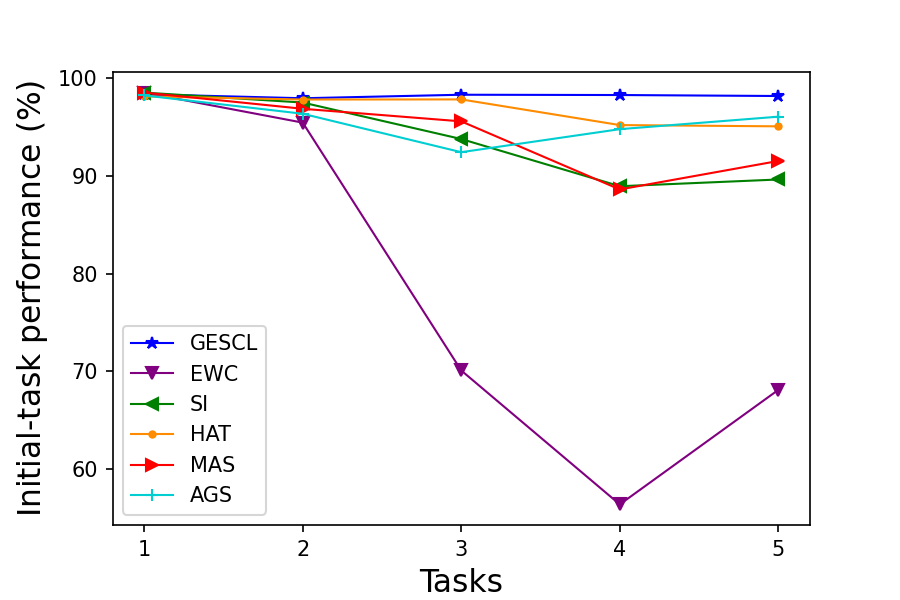}
        \caption{Split SVHN}
        \label{fig:exp1_svhn}
    \end{subfigure}
    ~ 
    \begin{subfigure}[b]{0.32\textwidth}
        \includegraphics[width=\textwidth]{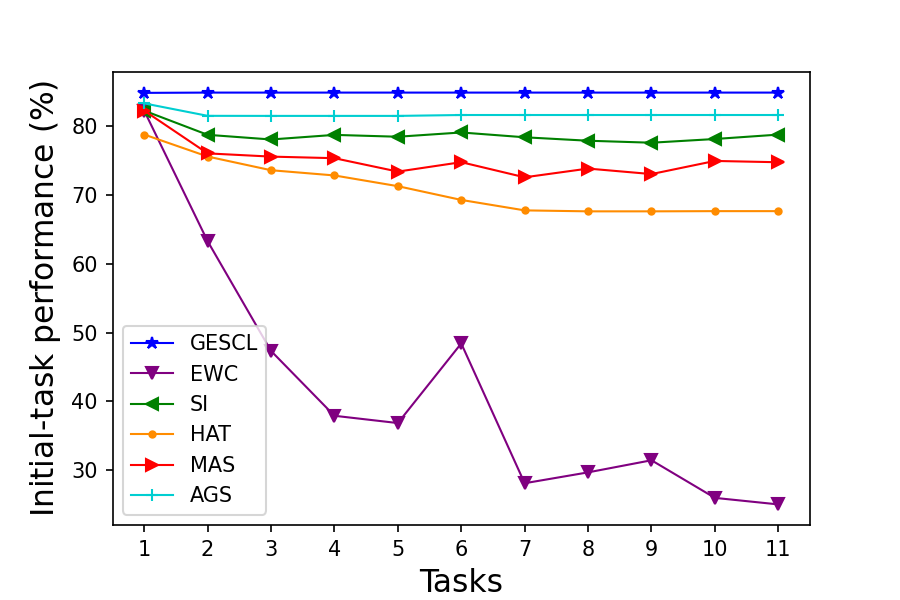}
        \caption{Split CIFAR-10/100}
        \label{fig:exp1_cifar10_100}
    \end{subfigure}
    ~
    \begin{subfigure}[b]{0.32\textwidth}
        \includegraphics[width=\textwidth]{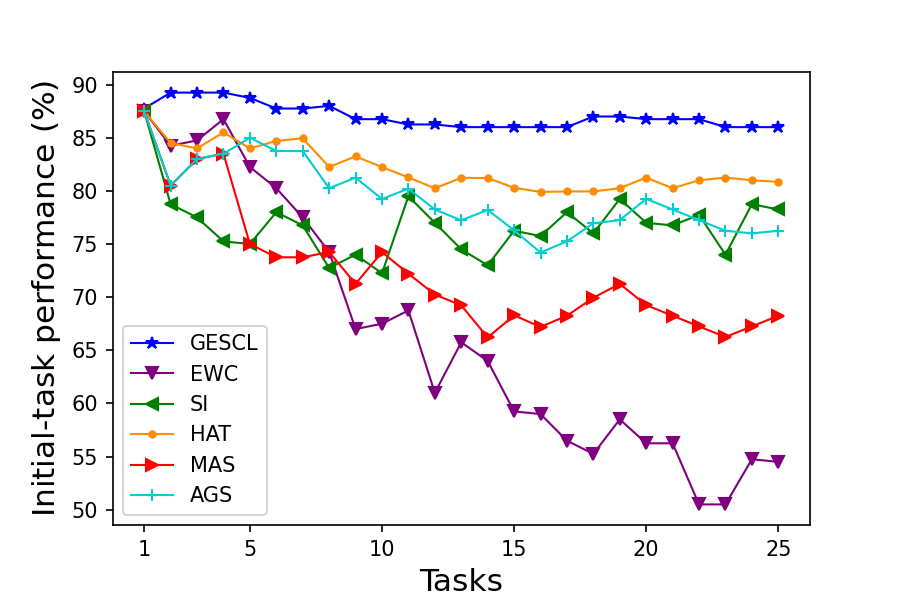}
        \caption{ImageNet-50}
        \label{fig:exp1_imagenet}
    \end{subfigure}
    \caption{Evaluation of catastrophic forgetting by measuring performance retention on the Initial task. Results show for each dataset how the classification accuracy of the first task evolves as further tasks are being learnt. Overall, GESCL show the strongest resilience against catastrophic forgetting. When facing new tasks, the performance it achieves on the initial task does not degrade as in concurrent approaches.}
    \label{fig:exp1_catastrophic_forg}
\end{figure*}

\section{Experiments}
\paragraph{\textbf{Continual Learning Scenario.}}
In the experiments, we follow the typical continual learning (CL) scenario as presented in Section~\ref{sec:intro}. To recall, in a CL scenario, a model is required to learn a sequence of tasks with unknown data distribution. We make the assumption that after training on task $t$, $D^t_{\text{train}}$ becomes inaccessible whereas $D^t_{\text{test}}$ is used to evaluate the model performance on $\mathcal{T}_t$ after training on $\mathcal{T}_t'$, with $t'> t$. Furthermore, we assume that the model capacity is fixed.

\paragraph{\textbf{Datasets.}} 
\vspace{0.4cm}
We conduct experiments on three different datasets, which are popular continual learning benchmarks: SVHN~\cite{netzer2011reading}, CIFAR-10/100~\cite{krizhevsky2009learning} and ImageNet-50\footnote{Our ImageNet-50 dataset can be downloaded \href{https://drive.google.com/file/d/145Z_zRlVMYAcxMPzqzh16W25LCYIdTOE/view?usp=sharing}{here}.}, which we generated as subset of the ImageNet dataset. It contains 50 classes grouped into 2 consecutive classes to form 25 tasks. Each class contains 1600 training images and 200 validation images.
For SVHN, which contain 10 classes, we group 2 consecutive classes to get 5 tasks. In the case of CIFAR-10/100, the first task consists of all 10 classes of CIFAR-10, wheras CIFAR-100 is split into 10 tasks, which serve as the remaining CL tasks, resulting in 11 tasks in this experiment. 

\paragraph{\textbf{Baselines.}}
\vspace{0.4cm}
We compare our work to prior state-of-the-art approaches including two reference methods EWC~\cite{kirkpatrick2017overcoming} and SI~\cite{zenke2017continual} as well as three recent, competitive ones, namely, MAS~\cite{aljundi2018memory}, HAT~\cite{serra2018overcoming} and AGS~\cite{jung2020continual}. 
We perform grid search to fairly select the best hyper-parameters for each approach. 

\paragraph{\textbf{Network.}}
\vspace{0.4cm}
For SVHN and CIFAR we use a CNN consisting of 3 blocks of $3 \times 3$ convolutions with $32, 64$ and $128$ filters respectively, followed by ReLU and a $2 \times 2$ max-pooling.
In ImageNet-50 experiments, we use a CNN similar to~\cite{vinyals2016matching}, which consists of two blocks of $2 \times 2$ convolution with $64$ filters, followed by ReLU and a $2 \times 2$ max-pooling. For all experiments, we used a multi-headed network.

\subsection{Is GESCL Able to not Catastrophically Forget?}
As an initial experiment, we examine the ability of our method at addressing catastrophic forgetting. To assess catastrophic forgetting, a common metric is to evaluate how the accuracy of each task varies when learning the remaining tasks~\cite{mirzadeh2020understanding}. 
This is shown in Fig.~\ref{fig:exp1_catastrophic_forg}, which illustrates for each dataset how the accuracy on the initial task evolves during the 

\begin{figure}[ht]
    \centering
    \begin{subfigure}[b]{0.32\textwidth}
        \includegraphics[width=\textwidth]{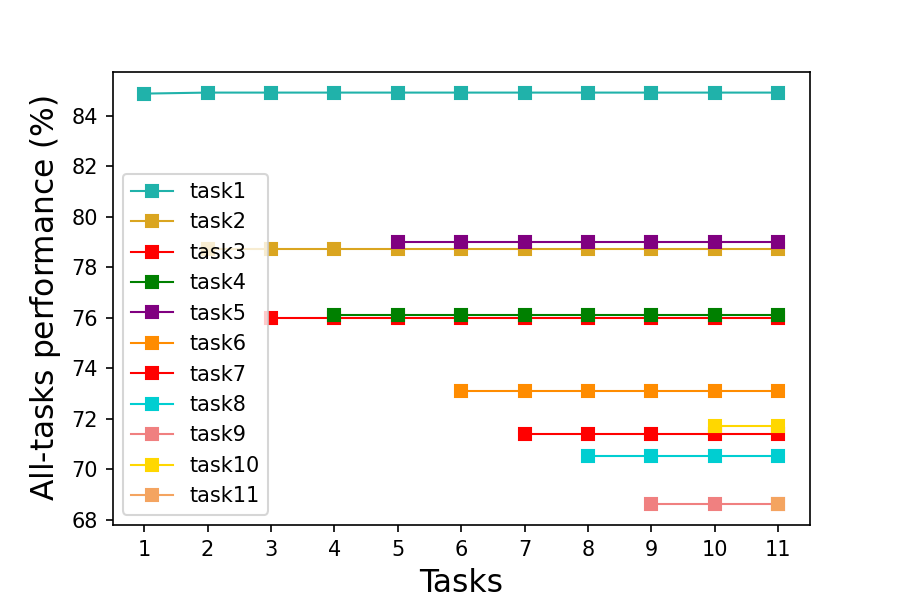}
        \caption{GESCL}
        \label{fig:exp2_gescl}
    \end{subfigure}
    ~ 
    \begin{subfigure}[b]{0.32\textwidth}
        \includegraphics[width=\textwidth]{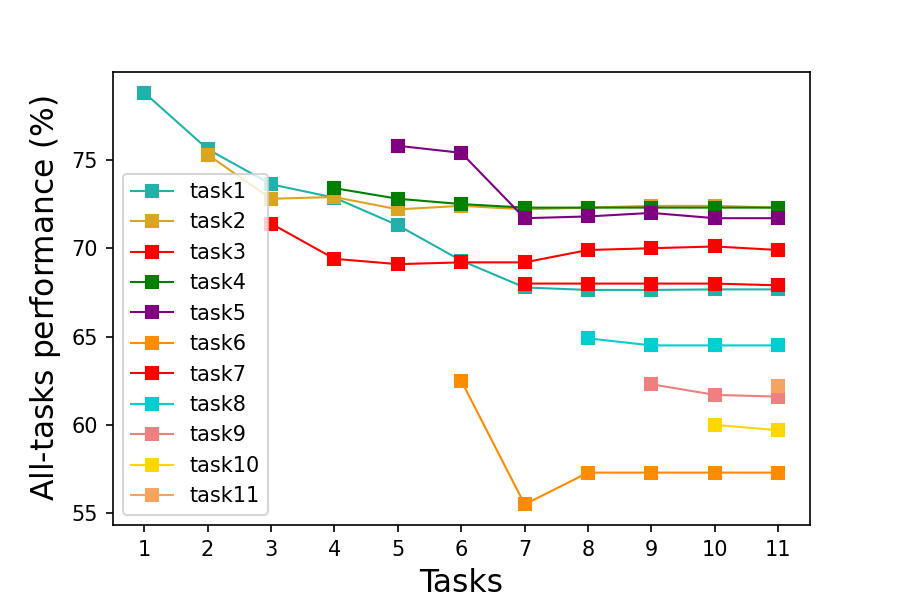}
        \caption{HAT}
        \label{fig:exp2_hat}
    \end{subfigure}
    ~
    \begin{subfigure}[b]{0.32\textwidth}
        \includegraphics[width=\textwidth]{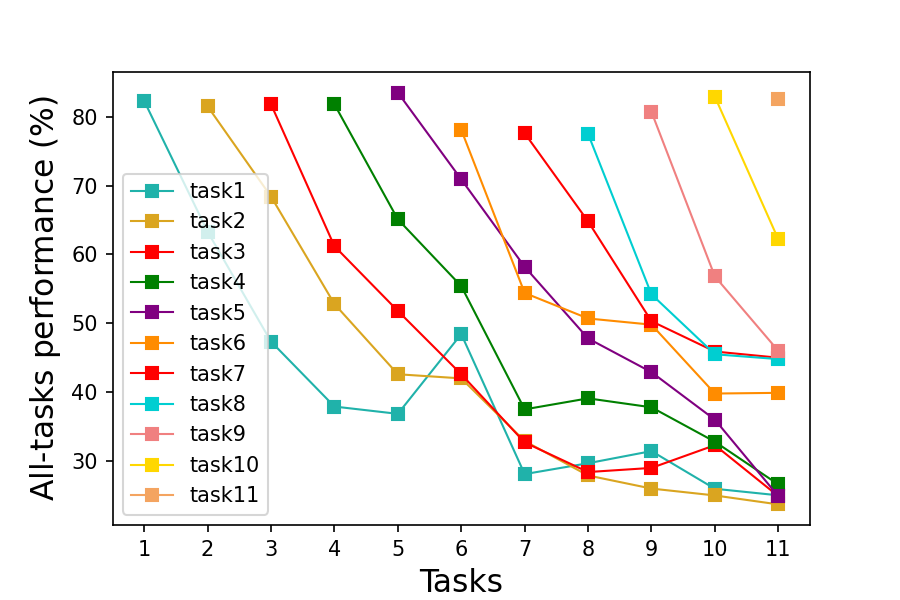}
        \caption{EWC}
        \label{fig:exp2_si}
    \end{subfigure}
    \caption{Evaluation of catastrophic forgetting by measuring performance retention on all task for the CIFAR-10/100 dataset. Results show how the test accuracy of each task evolves as further tasks are being learnt. Overall, GESCL suffers less from catastrophic forgetting.}
    \label{fig:exp2_catastrophic_forg}
\end{figure}

\begin{figure}[ht]
    \centering
    \begin{subfigure}[b]{0.32\textwidth}
        \includegraphics[width=\textwidth]{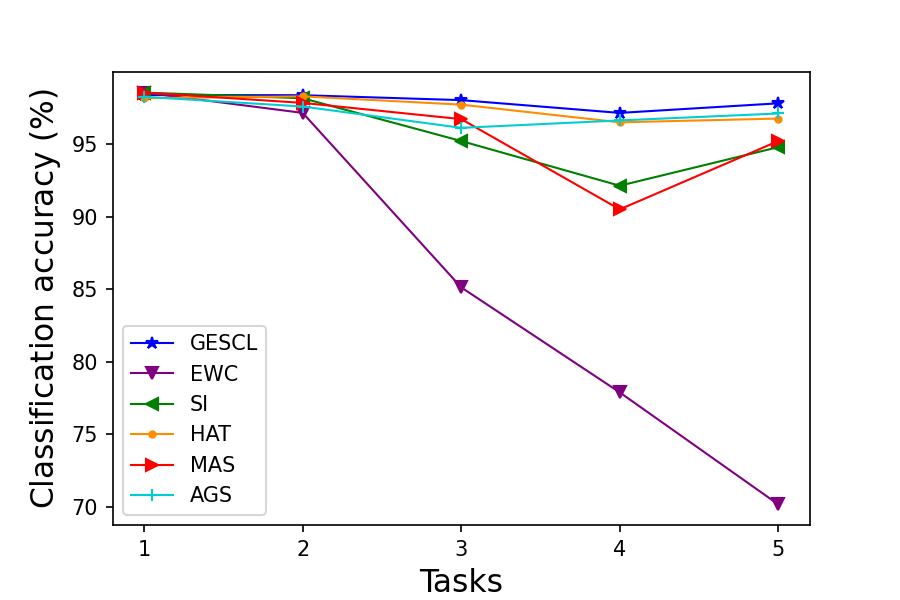}
        \caption{Split SVHN}
        \label{fig:exp3_svhn}
    \end{subfigure}
    ~
    \begin{subfigure}[b]{0.32\textwidth}
        \includegraphics[width=\textwidth]{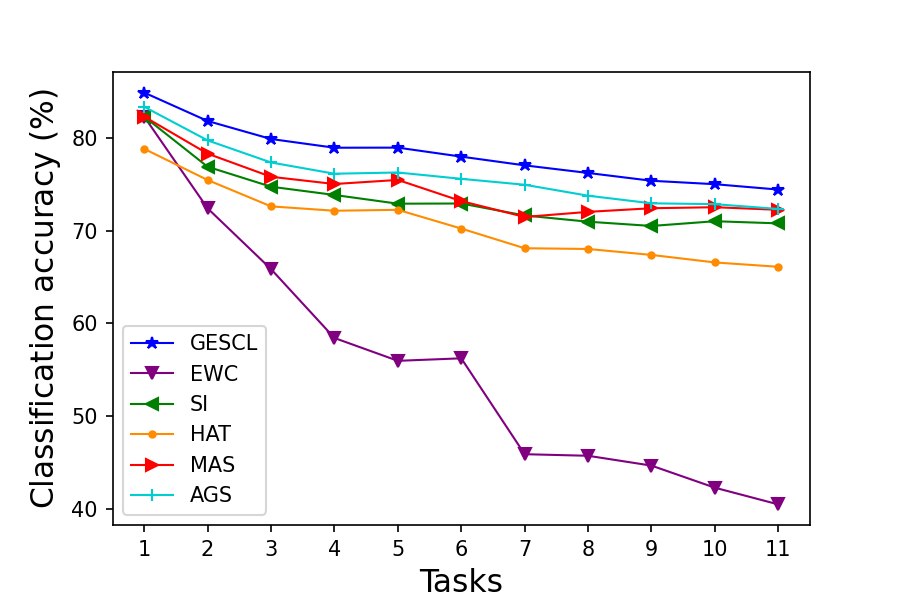}
        \caption{Split CIFAR-10/100}
        \label{fig:exp3_cifar}
    \end{subfigure}
    ~ 
    \begin{subfigure}[b]{0.32\textwidth}
        \includegraphics[width=\textwidth]{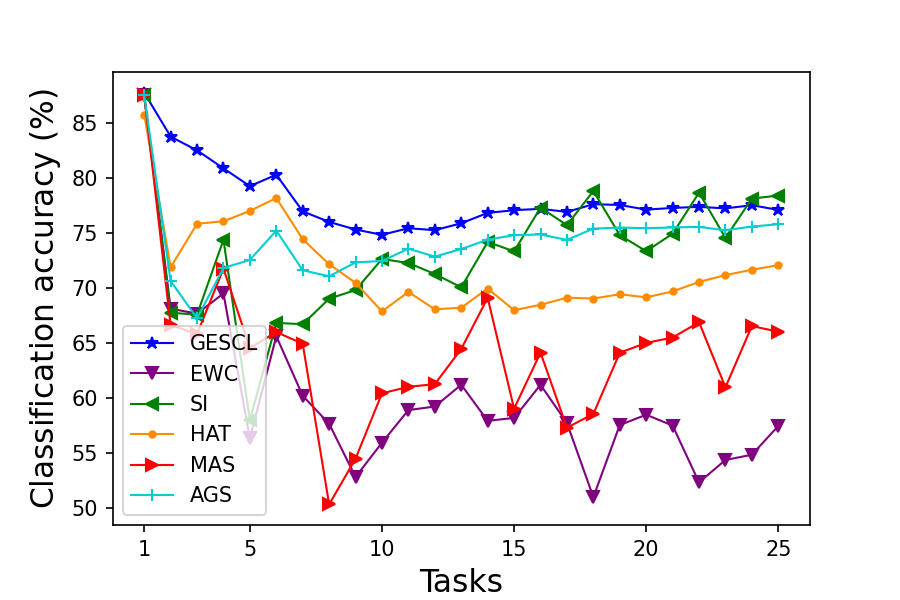}
        \caption{ImageNet-50}
        \label{fig:exp3_imagenet}
    \end{subfigure}
    \caption{Evaluation of average test  accuracy during the continual learning experiment of three datasets.}
    \label{fig:exp3_average_accur}
\end{figure}

\noindent
continual learning experiment. As can be seen, after sequentially training on all tasks, our method is the most stable and less forgetful, showing little to no forgetting on the ability to perform the first task.

To further examine the catastrophic forgetting evaluation, Fig.~\ref{fig:exp2_catastrophic_forg} outlines for the CIFAR-10/100 dataset, how the test accuracy of each task changes after new tasks have been learnt. 
Here again, GESCL shows strong resilience against catastrophic forgetting.

Overall, thanks to the filter importance adaptive stability regularizer (introduced in Section~\ref{sec:Methodoly}), which shrinks the change of vital parameters from task to task, \textsc{GESCL} outperforms state-of-the-art accuracy retention degrees. HAT~\cite{serra2018overcoming} tends to be a strong competitor especially on small dataset like SVHN, also achieving high performance retention degree, wheras EWC in contrast, suffer from severe degradation of performance especially in ImageNet-50 and CIFAR-10/100. This points out the limitation of EWC when the number of tasks becomes large.

\subsection{Is GESCL a Competitive Continual Learning Model in Terms of Accuracy?}
In this experiment, we aim at validating the effectiveness of our method in terms of classification accuracy during the continual learning experiments.
To evaluate the classification accuracy, we use the all-important average accuracy metric from~\cite{lopez2017gradient}, which is standard practice in the literature. More specifically, the classification accuracy at task $t$, is the average accuracy obtained from testing on task $1, \cdots, t$.
For instance, denoting by $a_t^{t'}$ the accuracy (fraction of correctly classified images) evaluated on the test set of the $t'$-th task after the model is sequentially trained on $\mathcal{T}_1$ to $\mathcal{T}_{t}$, the average accuracy $A_t \in [0,1]$ at task $t$ is:
\begin{equation}
    A_t = \frac{1}{t} \sum_{t'=1}^{t} a_t^{t'}.
    \label{eq:aver_accur}
\end{equation}
Fig.~\ref{fig:exp3_average_accur} provides for different datasets, a view of how the average accuracy evolves from task to task during the continual learning experiments. The results indicate that GESCL outperforms baseline algorithms. Interestingly, SI~\cite{zenke2017continual} reaches the best performance on the last task on ImageNet-50, after being clearly outperformed by our method on earlier tasks. Another interesting observation is that AGS~\cite{jung2020continual} is the strongest competitor to our method regarding average accuracy while it is surpassed by HAT~\cite{serra2018overcoming} on SVHN and ImageNet in terms of avoiding forgetting as illustrated in Fig.~\ref{fig:exp1_catastrophic_forg}. For both average accuracy and forgetting, our method shows superior performance to the aforementioned CL reference baselines.

An empirical conclusion that can be made out of Fig.~\ref{fig:exp1_catastrophic_forg},~\ref{fig:exp2_catastrophic_forg} and~\ref{fig:exp3_average_accur} is that GESCL achieves strong overall continual learning results, thanks to the way it addresses catastrophic forgetting and learning of several new tasks via the stability- and plasticity regularizers pair embedded in the networks learning procedure.

\subsection{Ablation Study}
We performed an ablation study to examine the contribution of each component of GESCL in its overall performance on CIFAR-10/100 and ImageNet-50 datasets. Particularly, we are interested in how (1) the stability regularizer $\mu_s$, (2) the plasticity regularizer $\mu_p$, (3) the filter importance balance $\nu$, contribute to the base model.  We implement variants of GESCL with different combinations of those components, each component being either activated ($\surd$) or removed ($\times$). We report the results in Table~\ref{tab:ablation_study}, where A denotes the average accuracy as defined in~Eq.~(\ref{eq:aver_accur}), and F indicates the average forgetting defined as the decrease in performance at each task between the peak accuracy and the accuracy after all tasks have been learnt~\cite{mirzadeh2020understanding}. Results for those two metrics as shown in Table~\ref{tab:ablation_study} demonstrate the effectiveness of  each component at improving the classification accuracy and overall forgetting of the method.

\begin{table}[ht]
    \centering
        \caption{Average accuracy (A) and average forgetting (F) of GESCL variants on CIFAR-10/100 and ImageNet-50.}
        \begin{tabular}{ccccccc} \\
        \toprule
        \multirow{2}{*}{$\mu_s$} & \multirow{2}{*}{$\mu_p$} & \multirow{2}{*}{$\nu$} & \multicolumn{2}{c}{CIFAR-10/100} & %
        \multicolumn{2}{c}{ImageNet-50} \\
        \cline{4-7}
         &&& A($\%$) & F & A($\%$)  & F \\
        \midrule
        \midrule
         $\surd$ & $\surd$ & $\surd$ & 74.5 & 0.0008 & 77.1 & 0.00007 \\
        \midrule
         $\surd$ & $\surd$ & $\times$ & 71.6 & 0.013 & 75.10 & 0.068 \\
         \midrule
        $\times$ & $\surd$& $\surd$& 41.9 & 0.44 & 57.38 & 0.349\\
         \midrule
         $\surd$& $\times$ & $\surd$ & 69.12 & 0.013 & 73.59 & 0.018\\
        \bottomrule \\
        \end{tabular}
    \label{tab:ablation_study}
\end{table}

\section{Conclusion}
In this work, we presented a simple yet effective continual learning method for convolutional neural networks. The proposed algorithm derived a novel regularization term to deal with the stability-plasticity dilemma. 
Through experiments, the algorithm showed consistent performance across standard continual learning benchmarks and performed competitive  or superior to existing state-of-the-art methods in both forgetting prevention and adaptability to new tasks.

\section*{Acknowledgements}
This work was funded by the federal state of North Rhine-Westphalia and the European Regional Development Fund FKZ: ERFE-040021.

\appendix



\printbibliography[heading=subbibintoc]

\end{document}

%% file: figs/filters.tex
\def\smallrect{3mm}
\begin{tikzpicture}[
    filter/.style={draw, minimum size=8mm, xslant=1}
]
\foreach \y in {-1, 0, 1} {
    \node [filter] (filter \y) at (0, \y) {};
}
\node [above=1ex of filter 1] {$w_i$};
\node [left=1ex of filter 1] {$h_i$};
\node [below=1ex of filter -1] {$X_i$};

\node [minimum width=4*\smallrect, minimum height=3*\smallrect, right=of filter 0] (tensor) {};
\fill [cyan] ([xshift=1*\smallrect]tensor.south west) rectangle ([xshift=2*\smallrect]tensor.north west);
\foreach \x in {0, ..., 4}
    \draw ([xshift=\x*\smallrect]tensor.north west) -- ([xshift=\x*\smallrect]tensor.south west);
\foreach \y in {0, ..., 3}
    \draw ([yshift=\y*\smallrect]tensor.south west) -- ([yshift=\y*\smallrect]tensor.south east);
\node [below=0 of tensor] {$c_{i+1}$};
\node [left=0 of tensor] {$c_i$};
\draw [stealth-] ([xshift=-0.5*\smallrect]tensor.north) -- ++(up:0.7)
      node [above] {$\mathcal{F}_{i,2}$};
\node [below=1.5em of tensor, align=left] {Kernel\\Tensor $\mathcal{F}_i$};

\foreach \y in {-3, -1, 3} {
    \node [filter] (filter \y) at (4, \y/2) {};
}
\node [filter, fill=cyan] (filter 1) at (4, 0.5) {};
\node [below=1ex of filter -3] {$X_o$};

\node [minimum width=5*\smallrect, minimum height=4*\smallrect] (tensor) at ($ (filter 1)!0.5!(filter -1) + (right:2.5) $) {};
\fill [cyan] ([xshift=3*\smallrect]tensor.south west) rectangle ([xshift=4*\smallrect]tensor.north west);
\fill [cyan] ([yshift=2*\smallrect]tensor.south west) rectangle ([yshift=3*\smallrect]tensor.south east);
\foreach \x in {0, ..., 5}
    \draw ([xshift=\x*\smallrect]tensor.north west) -- ([xshift=\x*\smallrect]tensor.south west);
\foreach \y in {0, ..., 4}
    \draw ([yshift=\y*\smallrect]tensor.south west) -- ([yshift=\y*\smallrect]tensor.south east);
\node [below=0 of tensor] {$c_{i+2}$};
\node [left=0 of tensor] {$c_{i+1}$};
\draw [stealth-] ([xshift=1*\smallrect]tensor.north) -- ++(up:0.7)
      node [above] {$\mathcal{F}_{i+1,4}$};
\end{tikzpicture}

%% file: references.bib
@incollection{mccloskey1989catastrophic,
  title={Catastrophic interference in connectionist networks: The sequential learning problem},
  author={McCloskey, Michael and Cohen, Neal J},
  booktitle={Psychology of learning and motivation},
  volume={24},
  pages={109--165},
  year={1989},
  publisher={Elsevier}
}

@article{abraham2005memory,
  title={Memory retention--the synaptic stability versus plasticity dilemma},
  author={Abraham, Wickliffe C and Robins, Anthony},
  journal={Trends in neurosciences},
  volume={28},
  number={2},
  pages={73--78},
  year={2005},
  publisher={Elsevier}
}

@inproceedings{yoon2017combined,
  title={Combined group and exclusive sparsity for deep neural networks},
  author={Yoon, Jaehong and Hwang, Sung Ju},
  booktitle={International Conference on Machine Learning},
  pages={3958--3966},
  year={2017}
}

@article{wen2016learning,
  title={Learning structured sparsity in deep neural networks},
  author={Wen, Wei and Wu, Chunpeng and Wang, Yandan and Chen, Yiran and Li, Hai},
  journal={arXiv preprint arXiv:1608.03665},
  year={2016}
}

@article{parikh2014proximal,
  title={Proximal algorithms},
  author={Parikh, Neal and Boyd, Stephen},
  journal={Foundations and Trends in optimization},
  volume={1},
  number={3},
  pages={127--239},
  year={2014},
  publisher={Now Publishers Inc. Hanover, MA, USA}
}

@article{jung2020continual,
  title={Continual Learning with Node-Importance based Adaptive Group Sparse Regularization},
  author={Jung, Sangwon and Ahn, Hongjoon and Cha, Sungmin and Moon, Taesup},
  journal={Advances in Neural Information Processing Systems},
  volume={33},
  year={2020}
}

@article{serra2018overcoming,
  title={Overcoming catastrophic forgetting with hard attention to the task},
  author={Serra, Joan and Suris, Didac and Miron, Marius and Karatzoglou, Alexandros},
  journal={arXiv preprint arXiv:1801.01423},
  year={2018}
}

@inproceedings{aljundi2018memory,
  title={Memory aware synapses: Learning what (not) to forget},
  author={Aljundi, Rahaf and Babiloni, Francesca and Elhoseiny, Mohamed and Rohrbach, Marcus and Tuytelaars, Tinne},
  booktitle={Proceedings of the European Conference on Computer Vision (ECCV)},
  pages={139--154},
  year={2018}
}

@article{kirkpatrick2017overcoming,
  title={Overcoming catastrophic forgetting in neural networks},
  author={Kirkpatrick, James and Pascanu, Razvan and Rabinowitz, Neil and Veness, Joel and Desjardins, Guillaume and Rusu, Andrei A and Milan, Kieran and Quan, John and Ramalho, Tiago and Grabska-Barwinska, Agnieszka and others},
  journal={Proceedings of the national academy of sciences},
  volume={114},
  number={13},
  pages={3521--3526},
  year={2017},
  publisher={National Acad Sciences}
}

@article{vinyals2016matching,
  title={Matching networks for one shot learning},
  author={Vinyals, Oriol and Blundell, Charles and Lillicrap, Timothy and Wierstra, Daan and others},
  journal={Advances in neural information processing systems},
  volume={29},
  pages={3630--3638},
  year={2016}
}

@article{liebenwein2019provable,
  title={Provable filter pruning for efficient neural networks},
  author={Liebenwein, Lucas and Baykal, Cenk and Lang, Harry and Feldman, Dan and Rus, Daniela},
  journal={arXiv preprint arXiv:1911.07412},
  year={2019}
}

@article{molchanov2016pruning,
  title={Pruning convolutional neural networks for resource efficient inference},
  author={Molchanov, Pavlo and Tyree, Stephen and Karras, Tero and Aila, Timo and Kautz, Jan},
  journal={arXiv preprint arXiv:1611.06440},
  year={2016}
}

@inproceedings{zhou2010exclusive,
  title={Exclusive lasso for multi-task feature selection},
  author={Zhou, Yang and Jin, Rong and Hoi, Steven Chu--Hong},
  booktitle={Proceedings of the thirteenth international conference on artificial intelligence and statistics},
  pages={988--995},
  year={2010},
  organization={JMLR Workshop and Conference Proceedings}
}

@inproceedings{zenke2017continual,
  title={Continual learning through synaptic intelligence},
  author={Zenke, Friedemann and Poole, Ben and Ganguli, Surya},
  booktitle={International Conference on Machine Learning},
  pages={3987--3995},
  year={2017},
  organization={PMLR}
}

@article{lopez2017gradient,
  title={Gradient episodic memory for continual learning},
  author={Lopez-Paz, David and Ranzato, Marc'Aurelio},
  journal={Advances in neural information processing systems},
  volume={30},
  year={2017}
}

@article{parisi2019continual,
  title={Continual lifelong learning with neural networks: A review},
  author={Parisi, German I and Kemker, Ronald and Part, Jose L and Kanan, Christopher and Wermter, Stefan},
  journal={Neural Networks},
  volume={113},
  pages={54--71},
  year={2019},
  publisher={Elsevier}
}

@article{krizhevsky2009learning,
  title={Learning multiple layers of features from tiny images},
  author={Krizhevsky, Alex and Hinton, Geoffrey and others},
  year={2009},
  publisher={Citeseer}
}

@article{netzer2011reading,
  title={Reading digits in natural images with unsupervised feature learning},
  author={Netzer, Yuval and Wang, Tao and Coates, Adam and Bissacco, Alessandro and Wu, Bo and Ng, Andrew Y},
  year={2011}
}

@article{mirzadeh2020understanding,
  title={Understanding the role of training regimes in continual learning},
  author={Mirzadeh, Seyed Iman and Farajtabar, Mehrdad and Pascanu, Razvan and Ghasemzadeh, Hassan},
  journal={arXiv preprint arXiv:2006.06958},
  year={2020}
}

@article{yoon2019scalable,
  title={Scalable and order-robust continual learning with additive parameter decomposition},
  author={Yoon, Jaehong and Kim, Saehoon and Yang, Eunho and Hwang, Sung Ju},
  journal={arXiv preprint arXiv:1902.09432},
  year={2019}
}

@article{von2021learning,
  title={Learning where to learn: Gradient sparsity in meta and continual learning},
  author={Von Oswald, Johannes and Zhao, Dominic and Kobayashi, Seijin and Schug, Simon and Caccia, Massimo and Zucchet, Nicolas and Sacramento, Jo{\~a}o},
  journal={Advances in Neural Information Processing Systems},
  volume={34},
  year={2021}
}

@article{ostapenko2021continual,
  title={Continual Learning via Local Module Composition},
  author={Ostapenko, Oleksiy and Rodriguez, Pau and Caccia, Massimo and Charlin, Laurent},
  journal={Advances in Neural Information Processing Systems},
  volume={34},
  year={2021}
}

@article{henning2021posterior,
  title={Posterior meta-replay for continual learning},
  author={Henning, Christian and Cervera, Maria and D'Angelo, Francesco and Von Oswald, Johannes and Traber, Regina and Ehret, Benjamin and Kobayashi, Seijin and Grewe, Benjamin F and Sacramento, Jo{\~a}o},
  journal={Advances in Neural Information Processing Systems},
  volume={34},
  year={2021}
}
